\documentclass[sigconf]{acmart}
\usepackage{multirow}
\usepackage{adjustbox}
\usepackage{makecell}
\usepackage{tablefootnote}
\usepackage{dsfont}
\usepackage{amsmath}
\AtBeginDocument{%
  }

\setcopyright{acmlicensed}
\copyrightyear{2024}
\acmYear{2024}
\acmDOI{XXXXXXX.XXXXXXX}

\acmISBN{\ }

\newcommand{\reffig}[1]{Fig. \ref{#1}}

\newcommand{\refeqn}[1]{Eqn. \ref{#1}}

\begin{document}

\title{Accelerating Inference of Networks in the Frequency Domain}


\author{Chenqiu Zhao}
\authornote{Chenqiu and Guanfang contributed equally to this work}
\affiliation{%
	\institution{Department of Computing Science, University of Alberta}
  \city{Edmonton}
  \country{Canada}}
\email{zhao.chenqiu@ualberta.ca}

\author{Guanfang Dong}
\authornotemark[1]
\affiliation{%
	\institution{Department of Computing Science, University of Alberta}
  \city{Edmonton}
  \country{Canada}}
\email{gfdong@@ualberta.ca}

\author{Anup Basu}
\affiliation{%
	\institution{Department of Computing Science, University of Alberta}
  \city{Edmonton}
  \country{Canada}}
\email{basu@ualberta.ca}

%
%
%
%



\begin{abstract}
It has been demonstrated that networks’ parameters can be significantly reduced in the frequency domain with a very small decrease in accuracy.
However, given the cost of frequency transforms, the computational complexity is not significantly decreased. 
In this work, we propose performing network inference in the frequency domain to speed up networks whose frequency parameters are sparse. 
In particular, we propose a frequency inference chain that is dual to the network inference in the spatial domain.
In order to handle the non-linear layers,
we make a compromise to apply non-linear operations on frequency data directly, which works effectively.
Enabled by the frequency inference chain and the strategy for non-linear layers,
the proposed approach completes the entire inference in the frequency domain.
Unlike previous approaches which require extra frequency or inverse transforms for all layers,
the proposed approach only needs the frequency transform and its inverse once at the beginning and once at the end of a network.
	Comparisons with state-of-the-art methods demonstrate that the proposed approach significantly improves accuracy in the case of a high speedup ratio (over 100x). The source code is available at \url{https://github.com/guanfangdong/FreqNet-Infer}.
\end{abstract}

%



\maketitle

\section{Introduction}
\label{sec_introduction}
Deep learning networks have achieved remarkable success in multiple fields, including multimedia, signal processing and computer vision.
However, the size of these networks and their computational cost limit their applications in multimedia, such as remote devices like cell phones.
Thus, accelerating network inference has become an interesting and important question.
Since frequency transforms are computationally expensive,
most recent methods focus on pruning networks for acceleration in the spatial domain rather than the frequency domain to demonstrate promising results \cite{Fang_2023_CVPR,gholami2022survey}.
However, it has been demonstrated that network parameters can be significantly truncated with the help of frequency regularization \cite{freReg}, demonstrating the potential for inference acceleration. 
In this work, we propose a novel method to accelerate network inference in the frequency domain. 
Unlike previous frequency domain methods that require extra frequency transforms for several layers, 
our work focuses on completing the entire inference process in the frequency domain. 
\begin{figure}[!t]
	\centering
	\includegraphics[width=\linewidth]{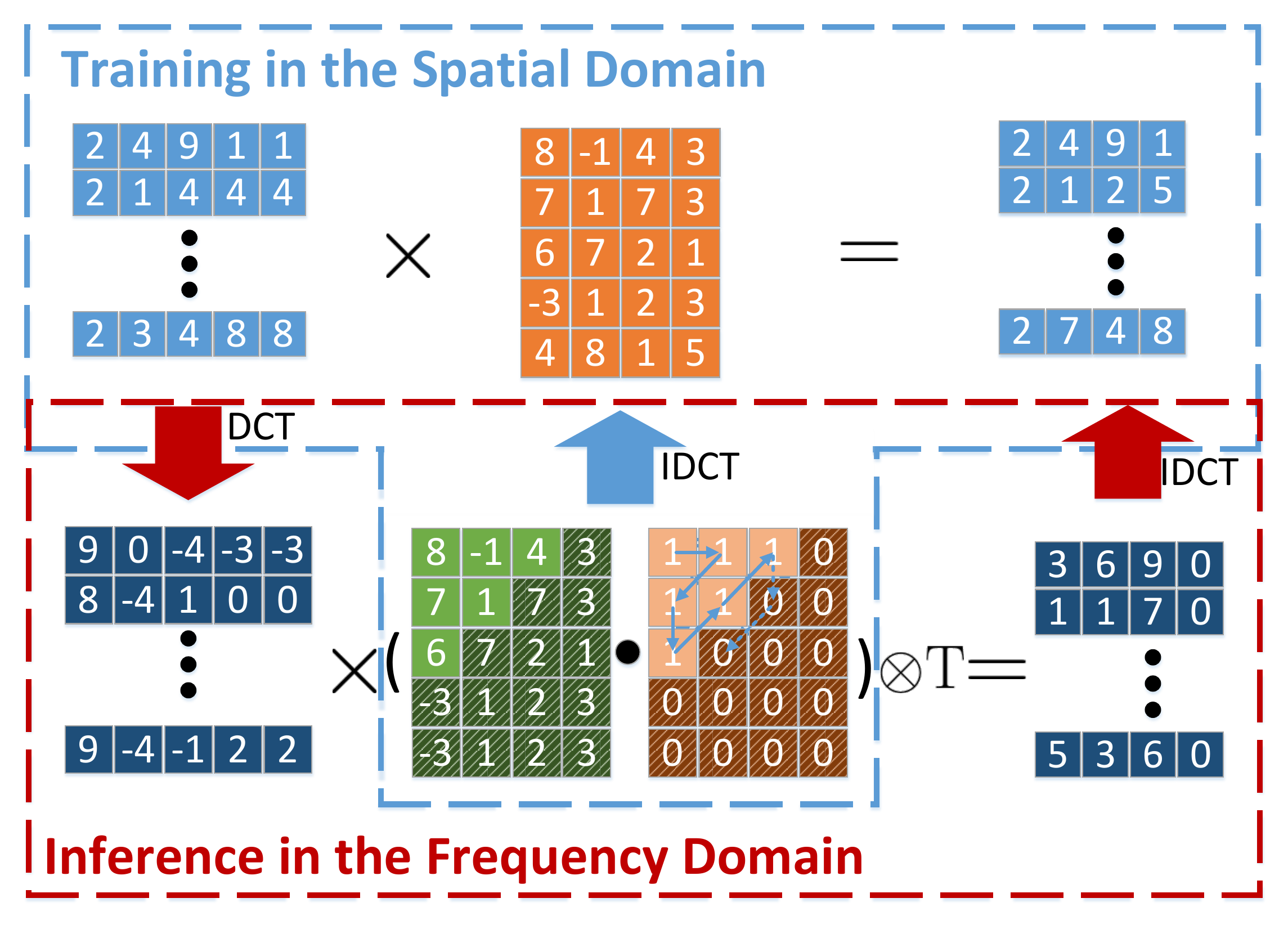}
	\caption{The proposed approach accelerates network inference in the frequency domain while keeping training in the spatial domain, to respect the fact that most of the network architectures are designed in the spatial domain.}
	\label{fig_idea}
\end{figure}

The main idea of the proposed approach is demonstrated in \reffig{fig_idea}.
We focus on accelerating the network inference in the frequency domain, while network training is done in the spatial domain.
The reason for training in the spatial domain is to respect the fact that most current architectures are designed in the spatial domain.
It would be very expensive to redesign these networks.
Since frequency regularization \cite{freReg} demonstrates good results in achieving high compression ratios, we believe it can accelerate networks even more effectively than pruning methods in the spatial domain.
This point is supported by our experiments, where the proposed approach shows significant accuracy advantages with a high speedup ratio between 10x to 150x compared to spatial methods.
Additionally, the existing frequency-domain pruning methods \cite{cheng2017survey} require extra frequency transformations and inverse frequency transformations between different layers, which is expensive and does not effectively accelerate the networks.
Therefore, there is a gap in the literature regarding accelerating inference in the frequency domain, without requiring extra frequency or inverse frequency transforms.

In this work, we propose a novel method to accelerate networks trained with frequency regularization which makes the parameters sparse in the frequency domain.
We propose a frequency inference chain that is dual to the inference in the spatial domain.
This dual correlation is extended from the convolution theorem, as demonstrated in \refeqn{eqn_fr_chain}.
With the help of the frequency inference chain, the \emph{entire inference} is performed in the frequency domain, requiring the discrete cosine transform (DCT) and inverse discrete cosine transform (IDCT) only once at the beginning and end of the network inference, as illustrated in \refeqn{eqn_infer}.
Another challenge is the non-linear layers, which break the inference chain in the frequency domain and lead to additional DCT and IDCT processes. 
To address this problem, non-linear layers are directly applied to the frequency data, maintaining the complete inference chain in the frequency domain.
These operations include layers such as Pooling, ReLU, and BatchNorm.
By resolving the challenges for both linear and non-linear layers, 
the proposed approach can perform the entire network inference in the frequency domain, achieving better results compared to state-of-the-art methods.
The main contributions of our work are:
\begin{itemize}
	\item We propose a frequency inference chain that is dual to matrix operations in the spatial domain. 
		The network can perform inference entirely in the frequency domain, requiring DCT and IDCT only once.
	\item We apply non-linear operations in the frequency domain. Although it is a compromise, it works effectively. This approach eliminates the need for DCT and IDCT processes when non-linear layers are present, and allows regular networks with non-linear layers to perform inference in the frequency domain.
	\item We perform comprehensive experiments to compare our method with state-of-the-art pruning methods. 
		Additionally, we have successfully built an application for mobile devices using the proposed method, demonstrating substantial savings in both inference speed and memory usage in a realistic setting. This application shows that our method has the potential to be universally deployed across a wide variety of multimedia devices.
\end{itemize}

\section{Related Work}
\label{sec_rel}
Methods accelerating network inference can be divided into quantization \cite{gholami2022survey, rokh2023comprehensive}, knowledge distillation \cite{gou2021knowledge, li2023object, chen2021distilling} and structured model pruning \cite{he2023structured, liang2021pruning}. 
Since structured model pruning changes the model structure, it can speed up the inference.
To achieve these goals, many algorithms are proposed in the spatial domain \cite{fang2023depgraph, li2017pruning, wang2020neural, liu2017learning, ma2023llm, fang2023structural, you2019gate, sekikawa2023bit, wangntk, wimmer2022interspace, sun2022disparse, redmanoperator, gao2024bilevelpruning, ilhan2024resource, ganjdanesh2024jointly, huang2024fedmef, lu2022learning, dao2021pixelated, he2017channel, he2018amc, he2019filter}. 
In general, spatial domain-based pruning methods aim to evaluate unimportant filters, channels, or blocks and remove them \cite{cheng2017survey}. For example, Fang et al. \cite{fang2023depgraph} constructs a dependency graph to explicitly model the dependencies and prunes the coupled parameter groups. He et al. \cite{he2018amc} propose a reinforcement learning method to automatically search the design space. Similarly, Ganjdanesh et al. \cite{ganjdanesh2024jointly} determine the pruning ratio based on reinforcement learning. Redman et al. \cite{redmanoperator} use the Koopman operator to analyze the dynamic behavior of the network for pruning. Gao et al. \cite{gao2024bilevelpruning} propose a unified dynamic and static channel pruning method. Ilhan et al. \cite{ilhan2024resource} prunes the model using importance estimation based on Taylor approximation. Huang et al. \cite{huang2024fedmef} proposes scaled activation pruning to save memory usage. These methods show that a network can be pruned and accelerated by spatially analyzing the framework.

Compared to spatial domain pruning, frequency domain pruning is less popular \cite{khaki2023cfdp, liu2018frequency, zhang2021filter, 2019_TPAMI_8413170, wang2016cnnpack, pmlr-v157-zeng21a, xu2020learning, gueguen2018faster, chen2020frequency}. Frequency domain pruning means the network is pruned in the frequency domain \cite{xu2019training}. Khaki and Luo \cite{khaki2023cfdp} proposed a frequency domain saliency metric to evaluate the importance of each channel. Liu et al. \cite{liu2018frequency} apply dynamic pruning to the DCT coefficients of network filters. Zhang et al. \cite{zhang2021filter} considered the input images and filter attributes, calculating the relevance measure in the frequency domain. Wang et al. \cite{2019_TPAMI_8413170, wang2016cnnpack} decomposed the representations of convolutional filters in the frequency domain into common parts and private parts. Zeng et al. \cite{pmlr-v157-zeng21a} obtained a compact model representation through low-rank approximation. Xu et al. \cite{xu2020learning} analyzed the spectral bias and demonstrated that CNN models are more sensitive to low-frequency channels. Gueguen et al. \cite{gueguen2018faster} reviewed JPEG and found faster and more accurate networks in the DCT space. Chen et al. \cite{chen2020frequency} simplified the convolution operations in the spatial domain to similar operations on frequency coefficients of input data and filters.

There are several important differences between the proposed approach and previous work based on frequency domain pruning.
Since the proposed approach completes inference in the frequency domain, including the input and the output,
the DCT and IDCT are only needed once at the beginning and once at end of the network.
In contrast, previous methods, such as \cite{2019_TPAMI_8413170}, \cite{pmlr-v157-zeng21a}, need DCT or IDCT for every convolution layer.
Note that it is also for of this reason, the proposed approach is able to truncate parameters on the first and last layers of networks, 
while state-of-the-art methods in the spatial domain, such as DegGraph \cite{Fang_2023_CVPR} or Slimming \cite{liu2017learning}, cannot.
In addition, almost all pruning methods in the frequency domain do not consider non-linear layers, such as ReLU or Norm, that are processed in the spatial domain, leading to extra DCT or IDCT processes.
As a result, the performance of previous methods in the frequency domain has almost no advantage compared to methods \cite{Fang_2023_CVPR, liu2017learning} in the spatial domain.
In the proposed approach, we made a compromise by applying non-linear operations in the frequency domain during training,
which addresses the problem of using many DCT and IDCT processes.
The proposed approach is thus able to achieve promising results compared with state-of-the-art pruning methods on a regular speedup ratio (10x),
and improvement in the case of a high speedup ratio (150x).
Also, compared to previous methods in the frequency domain,
we propose an application on a real mobile device and test the run time.
In contrast, previous approaches mainly present a theoretical speedup rate.

\section{Methodology}
\subsection{Preliminaries}
The proposed approach focuses on accelerating networks trained with frequency regularization, which is a recent technique to truncate parameters in the frequency domain.
For the completeness of this paper we briefly introduce the frequency regularization, please check \cite{freReg} for details.
Frequency regularization is proposed to restrict the non-zeros elements of the tensor of parameters in networks in the frequency domain.
In frequency regularization, tensors are maintained in the frequency domain in which element values on high-frequency locations are zigzag truncated during the training process.
Then, the truncated tensors are used as the input of the inverse discrete cosine transform to produce the spatial parameters that are used for regular training.
Assume the two N-dimensional tensors in the spatial domain and frequency domain to be $W(\vec{u}), \tilde{W}(\vec{v}) \in \mathbb{R}^{D_1 \times D_2, \cdots, D_N } $ where $\vec{u} = \{u_1, u_2, \cdots, u_N \} $ and $\vec{v} = \{v_1, v_2, \cdots, v_N \} $ are index vectors, $ u_i, v_i \in [0, D_i\!\!-\!\!1] \cap \mathbb{N}$.
Then, high dimensional frequency regularization is shown as:
\begin{equation}
	W(\vec{u}) = \text{IDCT}^{N} \left( \bigcup_{ \vec{x} \in G_{\vec{v}} } \tilde{W}(\vec{v}) \cdot \mathds{1}_{|\vec{v}|_1 < \epsilon}(\vec{v})  \right),
\end{equation}
where $\text{IDCT}^N$ is the N-dimensional inverse discrete cosine transform, which can be easily implemented by N IDCTs in different dimensions.
$\epsilon$ is the threshold value to control the truncation ratio. $|\vec{v}|_1 =  \sum_{i = 1}^{N}|v_i|$ is the $L_1$ norm of the index vector $|\vec{v}|$.
The indicator function $\mathds{1}_{|\vec{v}|_1 < \epsilon}(\vec{v})$ is used to approximate the zigzag binary mask for truncating parameters.
Benefiting from the learning ability of deep learning networks as well as the idea of gradually truncating parameters during training,
the networks trained with frequency regularization is very sparse in the frequency domain.
It demonstrates a great potential for network acceleration, which motivates the proposed approach in this work.

\subsection{Problem Statement}
Unfortunately, similar to most of the frequency pruning methods \cite{khaki2023cfdp, liu2018frequency, zhang2021filter, freReg}, networks trained with frequency regularization are only sparse in the frequency domain.
For network inference, data must be converted to the frequency domain, which requires extra DCT or IDCT processes.
For example, the PackCNN \cite{2019_TPAMI_8413170, NIPS2016_36366388}, which is one of the most popular frequency pruning methods, needs a partial DCT process for every convolution layer.
These extra DCT or IDCT processes limit the speedup ratio, which is the main reason why the frequency pruning methods like PackCNN \cite{2019_TPAMI_8413170} have no advantage compared to state-of-the-art pruning methods like DepGraph \cite{Fang_2023_CVPR} in the spatial domain.
To handle the cost of extra DCT or IDCT processes,
two challenges are:
\begin{itemize}
	\item Frequency Inference Chain: A regular convolutional layer output is still in the spatial domain, even when the sparse frequency kernels are involved in the computation.
		How to guarantee both the input and the output of every convolution layer are in the frequency domain is the first challenge we have to address.
	\item Non-linear Layers: When non-linear layers are involved in computation, data must be converted back to the spatial domain.
		This is not only because the non-linear layers are devised for spatial data, but also to keep the consistency of domains of data during training and testing, as the input of the output of non-linear layers are in the spatial domain during training.
\end{itemize}

\subsection{Frequency Inference Chain}
Our main focus is performing the entire inference process in the frequency domain, both the DCT and IDCT process are only done once.
In particular, DCT is applied to the input of the network, and IDCT is applied to the final output of the network.
To achieve this, we need to find a computation process in the frequency domain that is dual to the one performed in the spatial domain.
Assume $\mathbf{x}$ as the input of the convolutional neural networks.
The computation process of convolutional neural networks in the spatial domain can be presented as:
\begin{equation}
\mathcal{D}(\mathbf{x}) = (\cdots (\mathbf{x}W^{(1)} + B^{(1)})  \cdots)W^{(L)} + B^{(L)} =\prod_{\ell=1}^{L}\mathbf{x} W^{(\ell)} + R 
\end{equation}
where $\mathcal{D}(\cdot)$ denotes a neural network.
$\mathbf{x}$ is the input data. $W$ and $B$ represent the learning kernels and their basis.
$\ell$ is the layer index, $L$ is the total number of layers.
$R$ is a constant term contributed by all kernels in a network.
Assuming the network is trained with some frequency sparse technique,
the frequency tensor $\tilde{W}$ with respect to its spatial counterpart $W$ is very sparse.
Specifically,
many elements on the high frequency location of the $\tilde{W}$ is 0.
One reasonable method for network acceleration is the convolution theorem in which the spatial convolution is equal to the frequency dot product.
Since the original output of the convolution theorem is still in the spatial domain,
we propose a variation to make the output of a convolution still in the frequency domain which is shown as follows:
\begin{equation}
\text{DCT}(\mathbf{x}_{N \times M} \otimes W_{M \times K}) = \text{DCT}(\mathbf{x}_{N \times M}) \otimes \text{DCT}^{2}(W_{M \times K}) \otimes \text{T}, \\
	\label{eqn_fr_chain}
\end{equation}
where, $\text{DCT}(\cdot)$ is the discrete cosine transform and $\text{DCT}^2(\cdot)$ is the 2D discrete cosine transform. $\otimes$ is the matrix operation.
T is shown as follows:
\begin{equation}
\text{T} =  \text{D}^{-1}_{K\times K}(1,1) \odot \begin{bmatrix}
\frac{1}{2} & 0\\ 
0 &  I_{K-1 \times K-1}
\end{bmatrix} = \begin{bmatrix}
\frac{\text{D}^{-1}(1,1)}{2}  & 0\\ 
	0 &  I
\end{bmatrix},
\end{equation}
where $\text{D}^{-1}$ is the inverse basis matrix of $\text{DCT}(\cdot)$,
which is shown as follows:
\begin{equation}
\text{DCT}(\mathbf{x}) := \mathbf{x} \cdot \text{D}, \ \ \  \text{D}(i, j) = \text{cos}\left[\frac{\pi }{M}(i + \frac{1}{2})j \right ] \in R^{M\times M}.
\end{equation}
Note that some prior research \cite{2019_TPAMI_8413170, pmlr-v157-zeng21a} consider the basis matrix $\text{D}$ of DCT to be orthogonal, which is not entirely accurate. 
It would be more precise to state that the basis matrix of DCT-II is orthogonal after orthogonal normalization, whereas
the original DCT basis matrix is not orthogonal.
In our case, since the output of the convolution layer is devised to be in the frequency domain,
the orthogonal normalization is not applied for the DCT and IDCT basis matrix.
Instead, an extra term $\text{T}$ is attached to counteract the expansion in the results of the convolution theorem caused by the even extension in the DCT.
This is the main reason for the existence of T in \refeqn{eqn_fr_chain}.

With the help of \refeqn{eqn_fr_chain}, we can create the process that the input is the DCT output,
then all the involved kernels are in the frequency domain.
The equation becomes:
\begin{equation}
\begin{aligned}
\mathcal{D}(\mathbf{x}) & = \text{IDCT}(\text{DCT}(\mathcal{D}(\mathbf{x}))) = \text{IDCT}(\text{DCT}(\prod_{\ell=1}^{L}\mathbf{x} W^{(\ell)} + R )) \\
	& = \text{IDCT}(\prod_{\ell=1}^{L}\text{DCT}(\mathbf{x}) (\tilde{W}^{(\ell)} \text{T}^{(\ell)}) + \text{DCT}(R) )) \\
	& = \text{IDCT}(  \tilde{\mathbf{x}} \prod_{\ell=1}^{L} (\tilde{W}^{(\ell)} \odot \mathds{1}_{|\vec{v}|_1 < \epsilon}(\vec{v})) \text{T}^{(\ell)} + \tilde{R}) \\
	& = \text{IDCT}( \tilde{\mathcal{D}}( \tilde{\mathbf{x}})) 
\end{aligned}
\end{equation}
where $\tilde{\mathcal{D}}$ represents the frequency inference chain which is shown as follows:
\begin{equation}
	\begin{aligned}
	\tilde{\mathcal{D}}( \tilde{\mathbf{x}}) = & \prod_{\ell=1}^{L} \tilde{\mathbf{x}} \otimes (\tilde{W}^{(\ell)} \odot \mathds{1}_{|\vec{v}|_1 < \epsilon}(\vec{v})) \otimes \text{T}^{(\ell)} + \tilde{R}) \\
		= & (\cdots (\text{DCT}(\mathbf{x})\tilde{A}^{(1)} + \tilde{B}^{(1)})  \cdots)\tilde{A}^{(L)} + \tilde{B}^{(L)}
	\end{aligned}
	\label{eqn_infer}
\end{equation}
where $\tilde{A}$ is the tensor used for frequency inference which is shown as:
\begin{equation}
	\tilde{A} = (\tilde{W}^{(\ell)} \odot \mathds{1}_{|\vec{v}|_1 < \epsilon}(\vec{v})) \otimes \text{T}^{(\ell)}, \tilde{B} = \text{DCT}(B), 
\end{equation}
The proposed frequency inference chain is thus proposed in the \refeqn{eqn_infer}.
All the linear layers including the convolution layer, fully connected layer, and average pooling can be performed in the frequency domain, even if the training process is completed in the spatial domain.

\subsection{Non-linear Layer}
As we mentioned before,
the non-linear layers are handled by the compromise to directly apply frequency data for non-linear layers.
The original purpose of the non-linear layer is to lead to nonlinearity of the neural networks, rather than some computation requirements.
Moreover, inspired by the experiment in \cite{electronics10162004} where a ReLU layer is used for frequency data and the networks still work,
it seems to be possible to directly apply non-linear layers, as the neural network is believed to have excellent learning ability.
Thus, the frequency ReLU and BatchNorm are proposed:
\begin{equation}
\widetilde{\text{ReLU}}(\mathbf{x}) = \text{ReLU}(\text{DCT}(\tilde{\mathbf{x}})),
\end{equation}
\begin{equation}
\widetilde{\text{BatchNorm}}(\mathbf{x}) = \text{BatchNorm}(\text{DCT}(\tilde{\mathbf{x}})).
\end{equation}
During training, the input of ReLU and Batchnorm is applied to a DCT and then input into the regular ReLU and BatchNorm.
This is an arbitrary but useful idea. In our experiments, we only implemented ReLU, LeakReLU, and BatchNorm, which are enough for comparisons with state-of-the-art pruning methods.
However, the same idea can supposedly be applied to other non-linear layers.

Unfortunately, because ReLU and BatchNorm are originally designed for spatial data, its frequency forms slightly reduce the performance of network learning.
As a result, the accuracy of the proposed approach has around 3\% of accuracy decrease at low acceleration ratio (10x).
However, the proposed approach has significant improvement on a high acceleration ratio of around 150x.
Moreover, for other linear layers like pooling, frequency inference chains can be used directly.
Although we currently make a compromise to apply a non-linear layer such as ReLU for frequency data, it is possible to find an equivalent or approximate computation in the frequency domain as suggested by recent research \cite{10243466}.
This is because the Fourier transform can approximate any function.

\section{Experimental results}
\begin{table*}[th]
	\centering
	\begin{tabular}{|l@{\ }|@{\ }c@{\ }|c@{\ }|c@{\ }|c@{\ }|c@{\ }|c@{\ }|c@{\ }|c@{\ }|c@{\ }|c@{\ }|c@{\ }|c@{\ }|}
		\hline
		\makecell{Model} & \makecell{Speed\\Up} & \makecell{Li \\et al.\cite{li2017pruning}} & \makecell{Dep\\Graph\cite{fang2023depgraph}} & \makecell{Slimm\\ing\cite{liu2017learning}} & \makecell{GReg\cite{wang2020neural}} & \makecell{Proposed} & \makecell{Speed\\Up} & \makecell{Li \\et al.\cite{li2017pruning}} & \makecell{Dep\\Graph\cite{fang2023depgraph}} & \makecell{Slimm\\ing\cite{liu2017learning}} & \makecell{GReg\cite{wang2020neural}} & \makecell{Proposed} \\
		\hline
		Resnet18 
		& \multirow{8}{*}{10x} & 78.66\% & \textbf{91.34\%} & 80.16\% & 65.32\% & 89.82\% 
		& \multirow{8}{*}{50x} 
		& 67.44\% & 86.37\% & 77.09\% & 63.30\% & \textbf{88.67\%} \\
		Resnet34 & & 81.92\% & 89.05\% & 81.97\% & 67.81\% & \textbf{89.20\%} 
		& & 67.65\% & 87.60\% & 78.29\% & 64.84\% & \textbf{88.47\%} \\
		Resnet50 & & 83.68\% & 88.63\% & 82.15\% & 81.83\% & \textbf{89.48\%} 
		& & 79.86\% & 86.11\% & 79.20\% & 68.10\% & \textbf{89.37\%} \\
		Resnet56 & & 85.19\% & \textbf{88.09\%} & 83.56\% & 85.80\% & 85.32\% 
		& & 68.94\% & 76.07\% & 76.25\% & 70.94\% & \textbf{82.55\%} \\
		VGG8 & & 74.64\% & \textbf{88.77\%} & 68.76\% & 74.51\% & 84.93\% 
		& & 62.33\% & 78.58\% & 66.98\% & 68.79\% & \textbf{80.44\%} \\
		VGG13 & & 61.93\% & \textbf{91.91\%} & 63.69\% & 76.47\% & 90.17\% 
		& & 52.35\% & 74.80\% & 63.95\% & 72.45\% & \textbf{87.47\%} \\
		VGG16 & & 58.07\% & \textbf{92.41\%} & 51.05\% & 73.90\% & 88.70\% 
		& & 42.06\% & 80.26\% & 53.07\% & 74.53\% & \textbf{86.64\%} \\
		VGG19 & & 41.74\% & \textbf{92.29\%} & 51.29\% & 68.98\% & 87.94\% 
		& & 40.67\% & 83.03\% & 45.81\% & 63.04\% & \textbf{83.89\%} \\
		\hline
		Resnet18 & \multirow{8}{*}{100x} & 62.26\% & \textbf{85.61\%} & 61.49\% & 63.24\% & 85.35\%
		& \multirow{8}{*}{150x} 
		& 60.31\% & 76.26\% & 62.12\% & 60.53\% & \textbf{83.49\%} \\
		Resnet34 & & 66.03\% & 83.98\% & 76.56\% & 66.26\% & \textbf{85.80\%} 
		& & 66.04\% & \textbf{83.95\%} & 74.23\% & 63.07\% & 83.90\% \\
		Resnet50 & & 74.21\% & 79.95\% & 74.32\% & 58.38\% & \textbf{86.35\%} 
		& & 70.89\% & 71.32\% & 46.59\% & 53.10\% & \textbf{85.39\%} \\
		Resnet56 & & 68.41\% & 70.19\% & 68.28\% & 64.89\% & \textbf{73.69\%} 
		& & 58.79\% & 66.02\% & 33.69\% & 32.55\% & \textbf{70.49\%} \\
		VGG8 & & 60.86\% & 73.94\% & 55.62\% & 60.17\% & \textbf{76.39\%} 
		& & 54.33\% & 50.88\% & 49.90\% & 24.57\% & \textbf{76.12\%} \\
		VGG13 & & 49.44\% & 74.71\% & 54.17\% & 67.20\% & \textbf{81.74\%} 
		& & 47.81\% & 67.83\% & 44.04\% & 25.97\% & \textbf{81.65\%} \\
		VGG16 & & 43.38\% & 76.87\% & 44.75\% & 44.80\% & \textbf{82.27\%} 
		& & 25.07\% & 49.03\% & 50.74\% & 22.98\% & \textbf{82.62\%} \\
		VGG19 & & 25.02\% & \textbf{80.81\%} & 46.90\% & 44.61\% & 76.61\% 
		& & 22.28\% & 53.61\% & 21.40\% & 22.60\% & \textbf{75.74\%} \\
		\hline
	\end{tabular}
	\caption{Comparisons between the proposed approach and state-of-the-art methods on the CIFAR10 dataset.}
	\label{acceleration_comparison}
\end{table*}
\begin{figure*}[ht]
	\centering
	\includegraphics[width=0.99\linewidth]{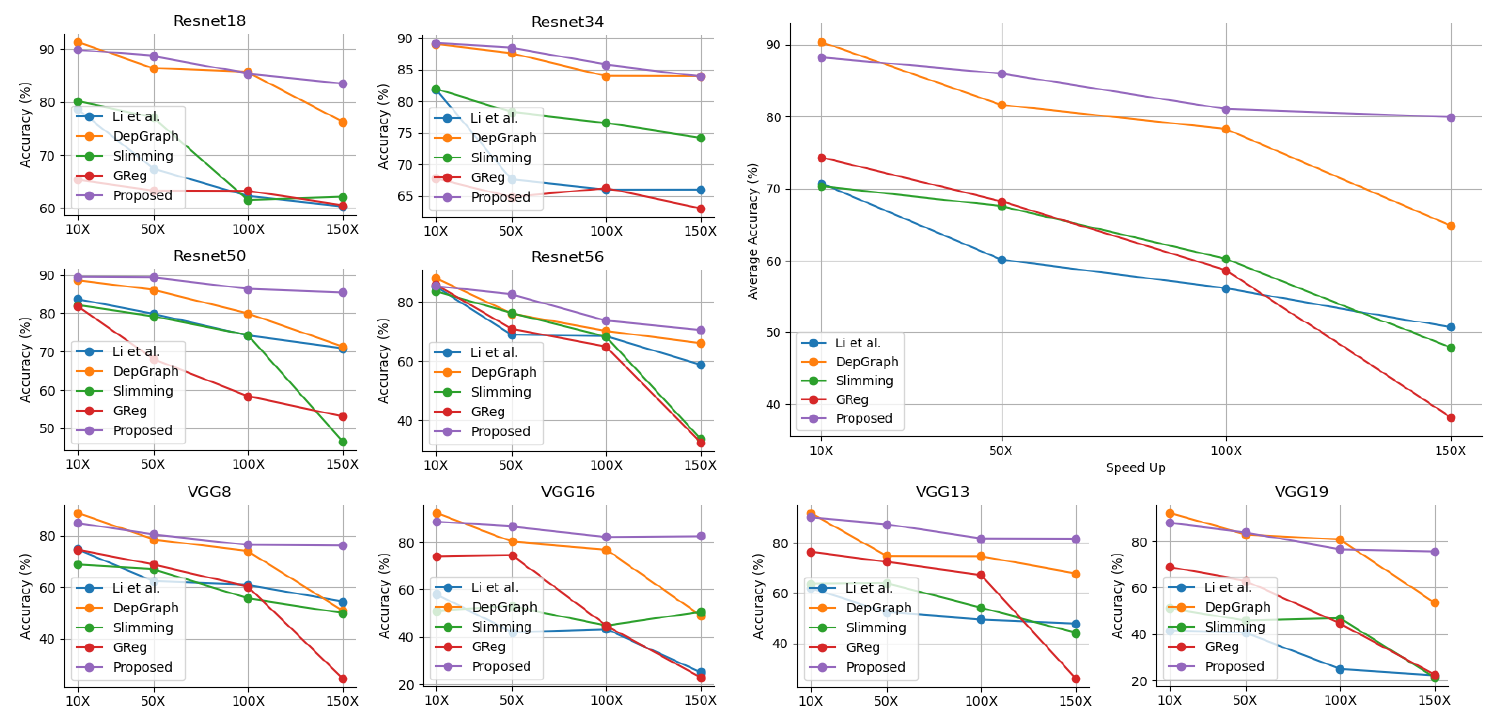}
	\caption{Comparison of different neural network speed up methods. The x-axis represents the inference speed-up rate, and the y-axis represents the classification accuracy in the figures. All experiments are conducted on the CIFAR-10 dataset. The top right figure shows the average accuracy across 8 network architectures, while the smaller figures illustrate the results for each model.}
	\label{fig:compare}
\end{figure*}

\subsection{Comparisons with State-of-the-art Methods}
\label{classification}
The proposed approach is evaluated on the CIFAR-10 dataset. 
To test the robustness of our method, we select popular neural network architectures, including ResNet-18, ResNet-34, ResNet-50, ResNet-56, VGG8, VGG13, VGG16, and VGG19.
To evaluate accuracy loss at various acceleration rates, we conduct experiments with speeds up to 10x, 50x, 100x, and 150x.
For comparisons, since most of the frequency pruning methods like PackCNN \cite{2019_TPAMI_8413170} did not demonstrate advantages compared to state-of-the-art spatial pruning methods,
the compared methods are mainly on pruning in the spatial domain, including DepGraph (CVPR23) \cite{fang2023depgraph}, GReg (ICLR21) \cite{wang2020neural}, Li et al. (ICLR17) \cite{li2017pruning}, and Slimming (ICCV17) \cite{liu2017learning}. 
We reproduce their results using the same acceleration rates and their publicly available code.
However, since our inference is performed in the frequency domain, our method is capable of pruning the input and output channels of the first and last layers, which non-frequency methods cannot achieve. Consequently, our actual acceleration rate is higher than that of the compared methods.

Table \ref{acceleration_comparison} presents the image classification results on the CIFAR-10 dataset with different speed-up rates and model configurations. Among the various methods, DepGraph is the best network acceleration method we could find. Our method outperforms it when the speed-up rate exceeds 10x. This demonstrates our method's superior ability to preserve accuracy at high speedup rates. 
The good performance of the proposed method at high speed-up rates is also verified in \reffig{fig:compare}. 
It clearly demonstrates the advantages of our methods for high speedup ratios around 150x.

Pruning methods in the spatial domain, like DepGraph, achieve network acceleration by analyzing the topological structure of the neural network. These methods model the dependencies of layers in network utilizing sparse learning.
Given the inherent learning redundancy of neural networks, methods like DepGraph typically perform well at lower speed-up rates, such as 10x. 
However, as the speedup rate increases, the networks's approximation ability decreases, which relies on the the number of remaining parameters.
%
%
This explains why the accuracy of networks that are pruned in the spatial domain has an obvious decrease for high speedup ratios.
For the proposed method, we take advantage of network inference in the frequency domain. By applying the discrete cosine transform (DCT) to the learning kernel of each layer, we can clearly distinguish between high-frequency and low-frequency signals.
This means that the parameters in the frequency domain are inherently assigned weights based on their importance. 
Consequently, we can prune parameters from the least important to the most important. By preserving the most critical parameters, our method maintains high accuracy even when the speed-up rate reaches 150x.

\subsection{Applications in Multimedia}
\subsubsection{Segmentation on U-Net}
\label{unet}
\begin{table}[h!]
	\centering
	\begin{tabular}{|c@{\ }|c@{\ }|@{\ }c@{\ }|@{\ }c@{\ }|@{\ }c@{\ }|@{\ }c@{\ }|}
		\hline
		\textbf{Model} & \textbf{Base} & \textbf{Freq. 1x} & \textbf{Freq. 2x} & \textbf{Freq. 10x} & \textbf{Freq. 100X} \\
		\hline
		\textbf{Dice Score} & 0.9855 & 0.9437 & 0.9313 & 0.9003 & 0.8801 \\
		\hline
	\end{tabular}
	\caption{Dice Scores for Different Models. The base model is the model without any modification. Freq. means the model is trained in the Frequency domain. 1x means the model has no speed up. 2X, 10Xand 100X means the model has 2, 10, and 100 times speed up.}
	\label{dice_score}
\end{table}
To test the generalization of the proposed method, we conduct experiments on segmentation. 
We apply a common U-Net to the Carvana Image Masking Challenge. 
The experimental results are shown in Table \ref{dice_score}. As we can see, our method maintains over a 0.9 dice score even with a 10 times speed up. 
However, when comparing only the Base model and the Freq. 1x model, we observe a 4\% accuracy decrease.
There are two reasons for this decrease: first is the computational error during frequency and inverse frequency transforms; and second is our strategy to handle the non-linear layers.
Since a non-linear layer like ReLU is designed for spatial data, directly applying it to frequency data results in a small decrease in accuracy. However, as we mentioned before, this accuracy cost is acceptable considering the reward at high speedup ratios.

\subsubsection{Applications on Mobile Devices}
\label{mobile}
Mobile devices, one of the most popular multimedia platforms, currently has a high demand for employing neural network inference. However, the deployment of deep learning models on mobile devices has two significant challenges: limited computational power and restricted memory capacity. Given that our method can accelerate inference speed while reducing memory consumption, it is meaningful to evaluate its performance on a mobile device.
\begin{table}[h!]
	\centering
	\renewcommand{\arraystretch}{1.4}
	\setlength{\tabcolsep}{2pt}
	\begin{adjustbox}{max width=\textwidth}
		\begin{tabular}{|l|r|r|r|c|c|c|}
			\hline
			\makecell{\textbf{Model}} & \makecell{\textbf{Batch} \\ \textbf{Size}} & \makecell{\textbf{Time}} & \makecell{\textbf{MAX} \\ \textbf{Memory}} & \makecell{\textbf{Layer}} & \makecell{\textbf{LeNet5} \\ \textbf{Shape}} & \makecell{\textbf{LeNet5*} \\ \textbf{Shape}} \\ \hline
			LeNet5 & \multirow{2}{*}{10000} & 1.1253s & 105.21MB & conv1 & (20, 1, 5, 5) & (2, 1, 5, 5) \\ \cline{1-1} \cline{3-7}
			LeNet5* &  & 0.1297s & 23.07MB & conv2 & (50, 20, 5, 5) & (5, 2, 5, 5) \\ \hline
			LeNet5 & \multirow{2}{*}{30000} & 3.3188s & 329.02MB & fc1 & (800, 500) & (80, 16) \\ \cline{1-1} \cline{3-7}
			LeNet5* &  & 0.4410s & 28.39MB & fc2 & (500, 300) & (16, 30) \\ \hline
			LeNet5 & \multirow{2}{*}{60000} & 7.735s & 679.17MB & fc3 & (300, 10) & (30, 2) \\ \cline{1-1} \cline{3-7}
			LeNet5* &  & 0.8242s & 46.96MB & \#Para. & 578,500 & 2,120 \\ \hline
		\end{tabular}
	\end{adjustbox}
	\caption{Performance comparison of LeNet5 and LeNet5* on a mobile device. LeNet5* is the accelerated network using our method.}
	\label{table:lenet5_comparison}
\end{table}

In this experiment, we employ the LeNet5 model for handwritten digit classification trained by the MNIST dataset. The test device is a Samsung Galaxy A54 5G (model SM-A546W), equipped with a Cortex A78 Quad-core 2.40 GHz CPU and 6 GB of RAM. We developed an application using Android Studio and the Gradle 8.0 plugin. Please check our demo video on YouTube (\url{https://youtu.be/CqkhKij6exI}).

Table \ref{table:lenet5_comparison} shows the real performance of deploying the proposed method on LeNet5. Since handwriting classification is a relatively easy task, we can significantly speed up the network. Theoretically, the accelerated LeNet5 requires 161,580 multiplications, while the original LeNet5 requires 20,441,000 multiplications, which is 126.5 times more than the accelerated one. In practice, our method achieves 8 to 10 times speedup in inference speed and 5-15 times memory savings. Smaller batch sizes save more time, while larger batch sizes save more memory. However, there is still a significant gap between the real performance and the theoretical speedup. We believe there are two reasons for this. First, the input data needs to be transformed to the frequency domain, which takes some time. However, this operation has a complexity of O(1), so it would not be a major issue for larger networks. Second, the real performance is affected by hardware/software optimization and I/O speed, which reduce the inference speed. Nonetheless, our method achieves remarkable performance.

\section{Spatial-frequency domain duality}
Currently, most state-of-the-art acceleration methods are proposed in the spatial domain. 
This preference is due to the computational cost of DCT and IDCT in the frequency domain. 
However, we address this issue by performing the entire inference in the frequency domain, necessitating only one DCT and one IDCT at the beginning and end of the network. 
This approach allows our method to achieve better results compared to state-of-the-art pruning methods in the spatial domain, despite previous frequency domain methods not showing any advantage. 
It is important to note that this does not imply that frequency domain methods are superior to spatial domain methods. 
Given the duality between spatial and frequency domains, the minimal information required for a network in either domain should be equivalent. 
Both spatial and frequency methods can enhance their performance with better sparsity algorithms, longer training times, or higher-precision GPUs. 
The primary difference lies in the training time or the speed to reach the minimal information, which is also the global optimal of training. 
Thus, both methods in the spatial domain or in the frequency domain have suitable applications. 
For instance, this paper demonstrates that frequency domain methods have an advantage at high compression rates for the same number of training epochs. 
Additionally, unlike spatial methods, frequency methods can drop the parameters in the first and last layers. 
Since frequency transforms are widely used in image and video compression, frequency networks can directly input compressed data, further accelerating computation speed. 
These are the advantages of frequency inference, and the main purpose of this paper is to highlight the potential of frequency pruning methods compared to spatial domain methods, thereby promoting the development of more frequency domain acceleration approaches.

\section{Conclusion}
We proposed a new method to accelerate the inference of networks trained with frequency regularization which makes the parameters of the networks sparse in the frequency domain.
In particular, the frequency inference chain is proposed to complete the inference process in the frequency domain. This strategy of applying a non-linear layer to frequency data guarantees the completeness of the frequency inference chain.
The proposed approach only requires DCT and IDCT once at the beginning and once at the end of the network,
whereas previous frequency domain methods usually require extra frequency or inverse frequency transforms for all layers.
The proposed approach achieves promising performance for high speedup ratios (around 100x). This is demonstrated by comprehensive experiments compared to state-of-the-art methods.
Finally, applications on mobile devices demonstrate the potential of our method to be universally deployed across a wide variety of multimedia devices.

\newpage

\bibliographystyle{IEEEtran}
\bibliography{ref}

\end{document}